\documentclass[12pt]{iopart}
\expandafter\let\csname equation*\endcsname\relax
\expandafter\let\csname endequation*\endcsname\relax 
\usepackage{amssymb,amsfonts, amsmath}
\usepackage{algorithmic}
\usepackage{graphicx}
\usepackage{textcomp}
\usepackage{xcolor}
\newcommand{\rb}{\right)}
\newcommand{\lb}{\left(}

\begin{document}

\title[Chaotic attractor reconstruction using small reservoirs]{
Chaotic attractor reconstruction using small reservoirs - the influence of topology
}

\author{Lina Jaurigue}

\address{Technische Universität Ilmenau}
\ead{lina.jaurigue@tu-ilmenau.de}
\vspace{10pt}

\begin{abstract}

Forecasting timeseries based upon measured data is needed in a wide range of applications and has been the subject of extensive research. A particularly challenging task is the forecasting of timeseries generated by chaotic dynamics. In recent years reservoir computing has been shown to be an effective method of forecasting chaotic dynamics and reconstructing chaotic attractors from data. 
In this work strides are made toward smaller and lower complexity reservoirs with the goal of improved hardware implementability and more reliable production of adequate surrogate models. 
We show that a reservoir of uncoupled nodes more reliably produces long term timeseries predictions than complex reservoir topologies. We then link the improved attractor reconstruction of the uncoupled reservoir with smaller spectral radii of the resulting surrogate systems. These results indicate that, the node degree plays an important role in determining whether the desired dynamics will be stable in the autonomous surrogate system which is attained via closed-loop operation of the trained reservoir. In terms of hardware implementability, uncoupled nodes would allow for greater freedom in the hardware architecture because no complex coupling setups are needed and because, for uncoupled nodes, the system response is equivalent for space and time multiplexing.

\end{abstract}

\section{Introduction}


Data driven models are needed in a wide range of applications and there are many statistical and machine learning methods of developing them. Low training costs and energy efficient approaches are paramount for wide spread use, particularly in edge devices. For data driven timeseries forecasting, particularly chaotic timeseries forecasting, reservoir computing has been shown to perform well. Reservoir computing is a machine learning approach wherein a dynamical system is driven with an input timeseries and the responses of the system to the input are linearly combined to approximate the desired output timeseries \cite{JAE01,MAA02}. In constrast to other machine learning methods, the training is restricted to the linear output weights and can be performed via linear regression. Due to the low training cost and the fact that the dynamical system remains unchanged in the training procedure, reservoir computing is well suited for hardware implementation. 

In this work strides are made toward smaller and lower complexity reservoirs which can lead to easier hardware implementation and interpretable surrogate models. We show that a reservoir of 20 uncoupled nodes more reliably produces better long term timeseries prediction performance than complex reservoir topologies. These results have implications for the hardware implementability, as uncoupled nodes are easier to construct and can equally be mutliplexed in space and time, giving more freedom for possible hardware architectures. Furthermore, to gain understanding about the connections between the reservoir topology and both the short- and long-term prediction performance, error measures which quantify the training accuracy, short-term predictions and the accuracy of the attractor reconstruction are analysed, as well as properties of the trained system. It is shown that an uncoupled reservoir topology leads to a smaller spectral radius of the final trained system and that this correlates with more stable attractor reconstruction.


The low training cost of reservoir computing has also lead to extensive theoretical and numerical studies. These have typically used larger numbers of reservoir nodes than the 20 used in this study. In \cite{PAT17} the closed loop operation of a reservoir trained on one-step-ahead prediction for the Lorenz and Kuramoto-Sivashinsky systems was investigated. The focus of this study was the estimation of Lyaponov exponents of the  Lorenz and Kuramoto-Sivashinsky systems and reservoirs of 300 nodes were used. In \cite{LU18} conditions for good short and long term prediction were established. The reservoir used in this study were sparse Erd\H{o}s-Renyi networks of 2000 nodes. In \cite{HAL19} the authors used a valid forecasting time measure, the correlation dimension and the largest Lyaponov exponent to quantify short and long term prediction quality. They investigate the influence of small world, Erd\H{o}s-Renyi and scale free network topologies, as well as the influence of the spectral radius of the original system. Networks of 300 nodes were used. 
In \cite{GRI19} various reservoir topologies were also compared for networks of 100 nodes, and Bayesian optimization was applied for hyperparameter optimisation. Optimisation and validation strategies for chaotic attractor reconstruction are investigated in \cite{RAC21}. In \cite{MA23} various block diagonal topologies were tested, with networks of 400-600 nodes. A block size of one is specifically not considered. The authors of \cite{MA23} compare the block diagonal topology with other reservoir computing related approaches in \cite{MA23a} and find good performance compared with the SINDy approach \cite{BRU16a}, nonlinear vector regression (referred to as "next generation reservoir computing" \cite{GAU21b}) and reservoir computing using a Erd\H{o}s-Renyi network. Performance comparisons of reservoir computing with other machine learning and statistical methods are given in \cite{GIL23}. Applications of trained chaotic attractors using reservoir computing have been investigated. For example, in \cite{ANT18a}
chaos synchronisation and cryptography applications are investigated and in \cite{ROE21} it is shown that such systems can be used to infer unseen attractors.

Several hybrid reservoir computing approaches for chaotic timeseries forecasting have been developed. One approach is to couple the reservoir at the output layer with a so-called knowledge-based model \cite{PAT18a}. This approach requires an approximate model for the underlying dynamics of the system to be reconstructed and has been shown to perform well, for example on atmospheric modelling \cite{ARC22}. Often an knowledge based model is not available, therefore in \cite{KOE23} a hybrid reservoir-SINDy approach was investigate. There the knowledge-based model was replaced with a model produced from data using the SINDy approach \cite{BRU16a}. 
Another important topic is the influence of input noise on the quality of the reconstruction of chaotic attractors. The authors of \cite{ZHA23e} have shown that injection noise can lead to improved short- and long-term predictions. And, in \cite{WIK24} the influence of input noise on the stability of reconstructed attractors, as well as the frequency components whereof, are investigated.

For many applications it is not only the reconstruction or forecasting of the dynamics that are relevant, it is often also desirable to gain insights into the underlying systems which are producing the data. This is difficult or even impossible with most machine learning approaches as they do not produce interpretable models.
One key step toward an interpretable model is to greatly restrict the size of the model. In terms of reservoir computing, this means restricting the dynamical degrees of freedom of the system used as the reservoir. A commonly used type of reservoir is a network of randomly connected nodes with nonlinear activation functions. For such a reservoir, the size of the reservoir is given by the number of nodes. In the context of interpretable models, the dynamics of the final trained system are of particular relevance, and these have so far not been thoroughly investigated.

In this work the influence of the reservoir topology on the ability to predict the short and long term behaviour of the Lorenz chaotic attractor is investigated, while restricting the number of reservoir nodes. The paper is structured as follows. In Section~\ref{sec:RC} the reservoir computing appproach is explain and the reservoir model and network topologies used in this work are introduced. In Section~\ref{sec:Task} the timeseries prediction task is presented and open- and closed-open predictions are explained. In Section~\ref{Sec:meth} various error measures are defined and simulation parameters are given. In Section~\ref{Sec:Results} the results are presented and discussed. Finally, the conclusions are given in Section~\ref{Sec:Conclusion}.


\section{Reservoir Computing}\label{sec:RC}

\begin{figure}[t]
\centerline{\includegraphics[width=0.9\textwidth]{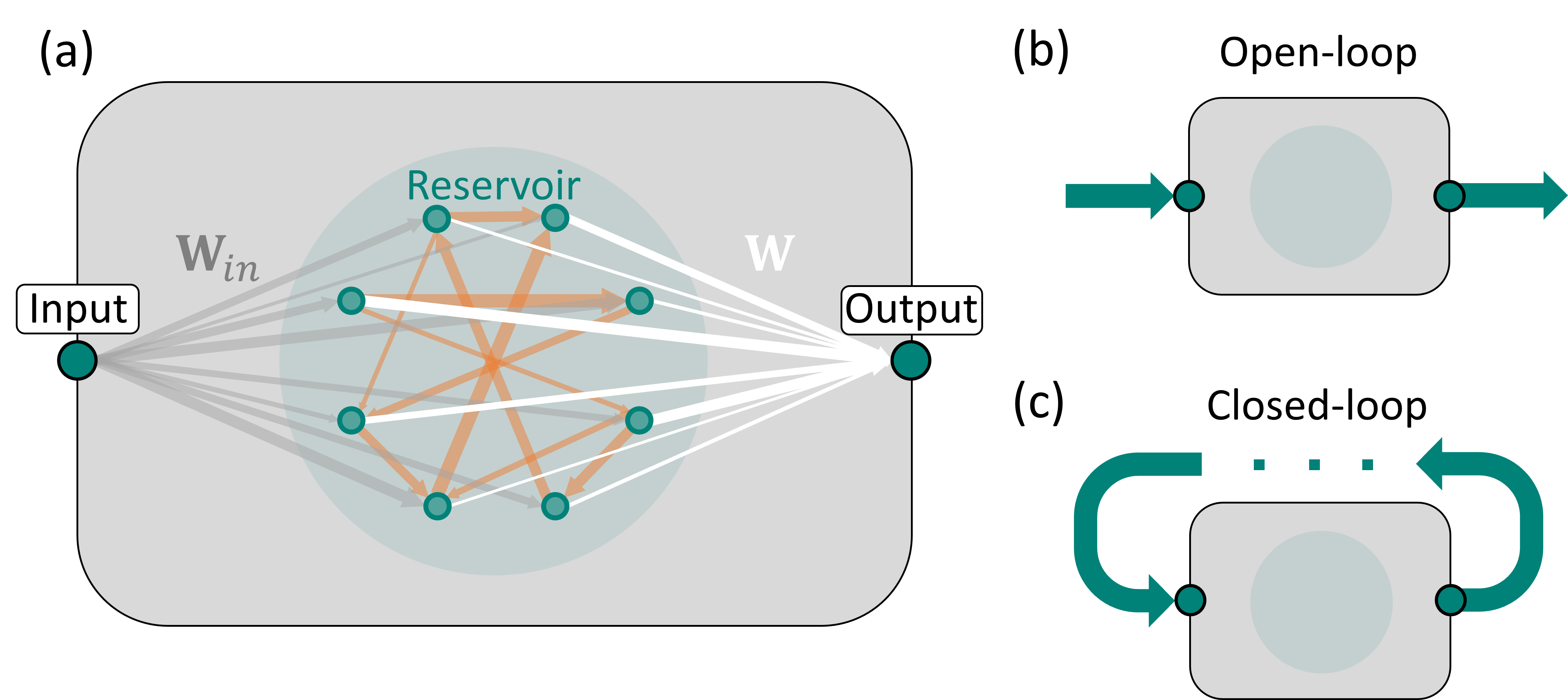}}
\caption{(a) Sketch of the reservoir computing approach. An input is fed into the reservoir with fixed random weights ${ \bf W}_{in}$ (grey arrows) and the output is produced by linearly combining states of the reservoir with trained output weights ${\bf W}$ (white arrows). (b) In open-loop operation externally sourced input data is fed into the reservoir to produce the output sequence. (c) In closed-loop operation the trained system is run autonomously by feeding the output back in as the next input step.}
\label{sketch_RC}
\end{figure}

In reservoir computing the response of a dynamical system to an input sequence is sampled a number of times and then a linear weighted sum of these sampled responses is used to make predictions or assign labels. In contrast to other machine learning approaches, the training is restricted to the linear output weights, as illustrated in Fig.~\ref{sketch_RC}a. Training the reservoir requires the following steps:
\begin{itemize}
    \item Feed a sequence of inputs of length $K_B+K_T$ into the reservoir.
    \item For each input step sample $N$ responses of the reservoir.
    \item Disregard the responses to the first $K_B$ inputs, where $K_B$ is long enough that the reservoir state no longer depends on the initial state.
    \item Collect the $N$ responses to the remaining $K_T$ inputs into a time ordered state matrix ${\bf S}\in\mathbb{R}^{K_T}\times \mathbb{R}^{N+1}$, adding a bias column of ones. The entry $s_{k,j}$ of the state matrix then corresponds to the $j^{\textrm{th}}$ sampled response of the reservoir to the $K_B+k^{\textrm{th}}$ input.
    \item Find the vector of weights ${\bf w} \in \mathbb{R}^{N+1}$ that minimises the difference between the target output ${\bf y}\in \mathbb{R}^{K_T}$ and the output ${\bf \hat{y}}={\bf 
S}{\bf w} \in \mathbb{R}^{K_T}$. The solution to this problem is given by 
\begin{equation}\label{Eq:wout}
    {\bf w}=({\bf S}^\textrm{T}{\bf S}+\lambda {\bf I})^{-1} {\bf S}^\textrm{T} {\bf y},
\end{equation}
using the Moore-Penrose pseudoinverse, where $\lambda$ is the Tikhonov regularisation parameter and ${\bf I} \in \mathbb{R}^{N+1}\times \mathbb{R}^{N+1}$ the identity matrix. For multiple targets ${\bf Y}=\left[{\bf y_1},..,{\bf y_q}\right]$ the matrix of output weights ${\bf W}=\left[{\bf w_1},..,{\bf w_q}\right] \in \mathbb{R}^{N+1}\times \mathbb{R}^{q} $ must be calculated.
    
\end{itemize}

\subsection{The reservoir}

Many different dynamical systems are suitable reservoirs and have been investigated both numerically and experimentally. These include optical \cite{BRU13a,LAR17,ROE19,SUG20,SKO22}, optoelectronic \cite{PAQ12,CHE19c}, micromechanical \cite{DIO18,YOS23}, quantum  reservoirs \cite{CHE20b,PFE22,MUJ23,CIN24}.
In this study networks of the following form are used as the reservoirs:
\begin{equation}\label{eq:ESN}
    {\bf x}\lb k+1\rb=\tanh\lb {\bf W}_{int}{\bf x}\lb k\rb +{\bf W}_{in}{\bf i}_{in}\lb k +1\rb\rb,
\end{equation}
where ${\bf x}\lb k\rb\in \mathbb{R}^M$ is the vector describing the state of the $M$ network nodes at discrete time $k$, ${\bf W}_{int}\in \mathbb{R}^{M}\times \mathbb{R}^{M}$ is the matrix of internal reservoir node couplings (adjacency matrix), ${\bf i}_{in}\lb k\rb \in \mathbb{R}^p$ is the vector of $p$ input signals at time $k$ and  ${\bf W}_{in}\in \mathbb{R}^M\times \mathbb{R}^{p}$ is the matrix of input weights. The function $\tanh\lb \cdot\rb$ is the nonlinear function which is applied element wise and commonly referred to as the activation function.
The $k^{\textrm{th}}$ row of the state matrix is filled by sampling $N$ of the reservoir states ${\bf x}\lb k\rb$, with $N \leq M$ and $k\in \left[0,K_T\right)$. For $N=M$ the elements of the state matrix are $s_{k,j}=x_j\lb k\rb$ with $x_j\lb k\rb$ ($j\in \left[0,..S\right)$) being the elements of the reservoir state vector ${\bf x}\lb k\rb$.

\begin{figure}[t]
\centerline{\includegraphics[width=0.6\textwidth]{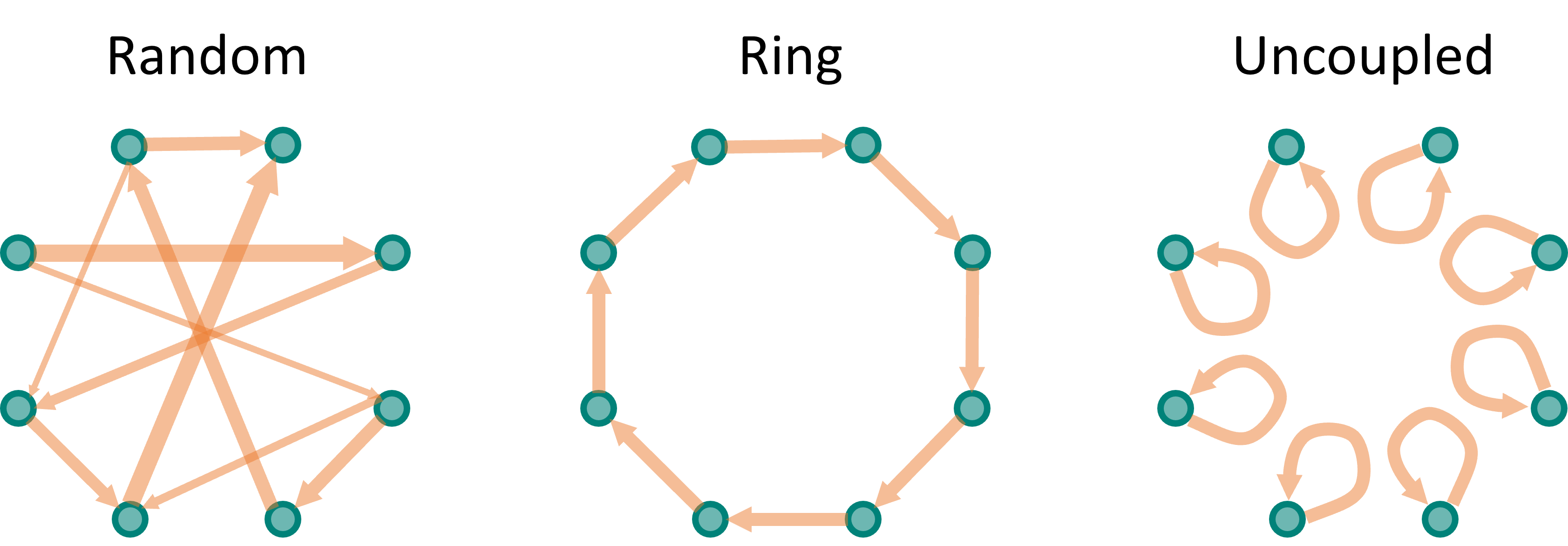}}
\caption{Reservoir network topologies corresponding to the coupling matrices ${\bf \tilde{W}}_{rand}$ (Random), ${\bf \tilde{W}}_{ring}$ (Ring) and ${\bf \tilde{W}}_{id}$ (Uncoupled).}
\label{sketch_topo}
\end{figure}

Three reservoir coupling matrix types are compared in this study; a random coupling matrix ${\bf \tilde{W}}_{rand}$ with entries chosen from a uniform distribution on the interval $[0,1]$, an identity matrix ${\bf \tilde{W}}_{id}$ and a diagonal matrix ${\bf \tilde{W}}_{ring}$ with the entries shifted one down from the main diagonal:
\begin{equation}
{\bf \tilde{W}}_{ring}=
    \begin{pmatrix}
        0 &0 &0&\cdots&0& 1\\
        1&0&0&\cdots&0 &0\\
        0& 1& 0&\cdots &0&0\\
        \vdots &\ddots&\ddots&\ddots &\ddots&\vdots\\
        0&0&0&\cdots&1&0
    \end{pmatrix}.
\end{equation}
In each case the matrices are rescaled such that ${\bf W}_{int}=\rho_R {\bf\tilde{W}}a_0$, where $a_0$ is the scaling factor needed such that $\rho_R$ is the spectral radius of ${\bf W}_{int}$. The spectral radius is defined as
\begin{equation} \label{eq:rho}
    \rho ({\bf W} ) =\max \{ |\lambda_1|, ..., |\lambda_N|  \},
\end{equation}
where $\lambda_1,...,\lambda_N$ are the eigenvalues of the $N\times N$ matrix ${\bf W}$.
For the ring and identity matrices $a_0=1$ and for the random matrix $a_0$ depends on the randomly drawn values.

For ${\bf W}_{int}={\bf W}_{id}$ the nodes are uncoupled in the training phase, i.e. each node only couples to its own state in the previous time step and the node degree is zero. This coupling scheme is identical to delay-based time-multiplexed reservoir computing when the feedback delay time  of the reservoir is equal to the input clocktime \cite{HUE22}. When ${\bf W}_{int}={\bf W}_{ring}$, each node is coupled to one neighbour, thereby forming a unidirectional ring. Sketches of the three coupling topologies are shown in Fig.~\ref{sketch_topo}.

If the coupling matrices are filled with zeros, then the reservoir reduces to an extreme learning machine (ELM), i.e. a single layer of uncoupled (also no self-coupling) nodes with random input weights \cite{HUA04}. An example of a hardware implemented ELM is presented in \cite{BIA23}, where an array of microresonators is used as a photonic ELM. In this limit, the system has no memory of past inputs and the method can also be interpreted as nonlinear vector regression where the feature vector contains various nonlinear transforms of the current input step. In the results section this limit is included and corresponds to the $\rho_R=0$ results (see Figs.~\ref{fig:nrmse}-\ref{fig:VPT2} in Section~\ref{Sec:Results}).

\section{Timeseries prediction}\label{sec:Task}

The aim of this study is to investigate the influence of the topology on the ability of the reservoir to predict chaotic dynamics and act as a surrogate model. To perform multi-step-ahead predictions and to act as a surrogate model the reservoir is trained to predict a timeseries one step ahead and then in the prediction phase the predicted output is used as the next input step. This is the so-called closed-loop operation and is illustrated in Fig.~\ref{sketch_RC}c (open-loop operation is shown in Fig.~\ref{sketch_RC}b). By coupling the reservoir output back in as the next input, the system becomes autonomous. The desired result would then be that the dynamics of the trained autonomous system emulate those of the original system which produced the training data. In this study the Lorenz system \cite{LOR63} is used to test to reservoir performance.

\subsection{Lorenz 63 system}\label{Sec:Lorenz}

The Lorenz system \cite{LOR63} is given by 
\begin{eqnarray}
\frac{dx}{dt}=c_1y-c_1x,\quad
\frac{dy}{dt}=x(c_2-z)-y, \quad \textrm{and} \quad
\frac{dz}{dt}=xy-c_3z.\label{Lorenz}
\end{eqnarray}
With $c_1=10$, $c_2=28$ and $c_3=8/3$ this system exhibits chaotic dynamics and has a Lyaponov exponent of $\lambda_L\approx 0.91$. We generate the Lorenz timeseries using Runge-Kutta fourth order numerical integration with a time step of $h=10^{-3}$. We use all three variables, sampled with a step size of $\Delta t=0.1$ and rescaled to the range $\left[0,1\right]$, as the reservoir input ${\bf i}_{in}(k)=\left[X(k),Y(k),Z(k)\right]$ ($X$, $Y$, $Z$ indicate the rescaled versions of $x$, $y$ ,$z$). In the training phase the target is given by ${\bf y}(k)=\left[X(k+1),Y(k+1),Z(k+1)\right]$. The step size of $\Delta t=0.1$ is chosen, as it has been shown to be well suited for the reconstruction of the Lorenz attractor \cite{KAN03,TSU23}.

\subsection{The trained autonomous system}

After training, the reservoir output can be written in terms of the reservoir state:
\begin{equation}
{\bf \hat{y}}\left(k\right) ={\bf x}(k)^\textrm{T}{\bf W},
\end{equation}
where the reservoir state vector is extended by one term, $x_{N+1}=1$, to include the bias. 
Labelling the entries of the output weight matrix (${\bf W}$) which correspond to the bias terms as $W_{N+1,i}=W{^\textrm{bias}_i}$, the reservoir outputs can be written as
\begin{equation}
\hat{y}_i(k)=\sum_{j=1}^{N}x_j(k)W_j^i+W{^\textrm{bias}_i},
\end{equation}
for $i\in \left[1,q\right]$, where $q$ is the number of target sequences.
Then the next input is given by $I_i(k+1)=\hat{y}_i(k)$, i.e. ${\bf i}_{in}(k+1)=\left[\hat{y}_x(k),\hat{y}_y(k),\hat{y}_z(k)\right]$

The reservoir then becomes an autonomous dynamical system described by 
\begin{equation}\label{eq:ESNclosed}
    {\bf x}\lb k+1\rb=\tanh\lb {\bf W}^a_{int}{\bf x}\lb k\rb +{\bf b}\rb,
\end{equation}
where ${\bf W}^a_{int}$ is the modified coupling matrix and ${\bf b}$ is the vector of constant bias terms:
\begin{equation}
    {\bf b} =\sum_{i=1}^q {\bf w}_{in}^i W_{\textrm{bias}}^i,
\end{equation} 
with ${\bf W}_{in}=[{\bf w}_{in}^1,...,{\bf w}_{in}^q]$.
The righthand side of the final system now only dynamically depends on the previous states of the nodes. 

For the case of reconstructing a three dimensional dynamical system ($q=3$) the new coupling matrix is given by
\begin{equation}
{\bf W}^a_{int}={\bf W}_{int}+
\begin{pmatrix}
\sum_{i=1}^3 W_{in,1}^iW_1^i & \sum_{i=1}^3 W_{in,1}^iW_2^i & \cdots & \sum_{i=1}^3 W_{in,1}^iW_N^i\\
\sum_{i=1}^3 W_{in,2}^iW_1^i & \sum_{i=1}^3 W_{in,2}^iW_2^i & \cdots &\sum_{i=1}^3 W_{in,2}^iW_N^i\\
\vdots&\vdots&\vdots&\vdots\\
\sum_{i=1}^3 W_{in,N}^iW_1^i&\sum_{i=1}^3 W_{in,N}^iW_2^i& \cdots&\sum_{i=1}^3 W_{in,N}^iW_N^i
\end{pmatrix},
\end{equation}
the entries of which depend on the initial reservoir, the input weights and the trained output weights. We will relate spectral radius of this new coupling matrix to the ability of the reservoir to operate as a surrogate model. The spectral radius of ${\bf W}^a_{int}$ is calculated using Eq.~\eqref{eq:rho} and will be denoted $\rho_a$.

\section{Methods}\label{Sec:meth}

\subsection{Performance measures}

\subsubsection{Normalised root mean squared error}

The performance of the open-loop (see Fig.~\ref{sketch_RC}b) one-step-ahead prediction is quantified using the normalised root mean squared error (NRMSE), defined as
\begin{equation}
 \textrm{NRMSE}=\sqrt{\frac{\sum_{k=1}^{K_T}\lb y_{k}-\hat{y}_{k}\rb^2}{K_T \textrm{var}\lb {\bf y}\rb}},
\end{equation}
where $y_k$ are the target values, $\hat{y}_{k}$ are the outputs produced by the reservoir computer, $K_T$ is the number of testing steps (i.e. the length of the vector ${\bf y}$) and $\textrm{var}\lb {\bf y} \rb$ is the variance of the target sequence.

\subsubsection{Valid Prediction Time}

We quantify the short-term closed-loop (see Fig.~\ref{sketch_RC}c) performance using a valid prediction time (VPT) as in \cite{KOE23}. The valid prediction time is given by 
\begin{equation}
    t_{\textrm{VPT}}=\textrm{max}\{ t |\delta_{\textrm{VPT}}<0.4\},
\end{equation}
where $t$ is the time determined by the discretisation step $\Delta t$ of the training data set and number of iterations from the beginning of the prediction phase, and $\delta_{\textrm{VPT}}$ is a measure of the prediction error defined as:
\begin{equation}
    \delta_\textrm{VPT}=\frac{|{\bf y}(k)-{\bf \hat{y}}(k)|^2}{\langle|{\bf y}(k)-\langle {\bf y}(k)\rangle|^2\rangle}.
\end{equation}
Here $\langle \cdot\rangle$ indicates the time average over the length of the target timeseries, ${\bf y}(k)=\left[X(k+1),Y(k+1),Z(k+1)\right]$ is the vector of target values at discrete time $k$ and ${\bf \hat{y}}(k)=\left[\hat{X}(k+1),\hat{Y}(k+1),\hat{Z}(k+1)\right]$ is the vector of predicted values.

Various other definitions of prediction times are used in the literature. For example in \cite{MA23} a forecast horizon is defined as the time at which the difference in the predicted and target trajectories exceeds the standard deviation of the target trajectory. Care must be taken when comparing these measures as they can lead to very different times.

\subsubsection{Attractor Deviation}\label{sec:adev}

To quantify how well the desired attractor is reconstructed we use an approach similar to that used in \cite{ZHA23e}. We divided the 3-dimensional phase-space into cubes of volume $dX\times dY\times dZ$. Each cube is assigned a value of one or zero; zero if the trajectory never passes through the cube and one if the trajectory passed through at least once. We do this for both the reconstructed trajectory and the target trajectory. We then add together the absolute value of the difference of the values assigned to each cube for the reconstructed and target trajectories:
\begin{equation}
\textrm{ADev}=\sum_i\sum_j\sum_k| (c^{true}_{ijk}-c^{pred}_{ijk})|,
\end{equation}
where $c^{true(pred)}_{ijk}$ is one if the trajectory passed through the corresponding element of the phase space at least once and otherwise zero.
The ADev value depends on the size of the cubes, on the length of the timeseries and on the target trajectory. We choose a cube dimension of $0.1\times 0.1\times 0.1$ and a timeseries length of 5000 time units ($5000\Delta t$), which is about 450 Lyaponov times ($1/\lambda_L\approx 1.1$) for the Lorenz 63 system with the standard parameters used here. 

\subsubsection{Power Spectral Density}

To compare the frequency components of the target and reconstructed trajectories we calculate the power spectral density (PSD) of the $Z$ component. This is done using the timeseries over the prediciton horizon of $5000\Delta t$, applying a Hamming windowing function and then calculating the fast Fourier transform (FFT). The PSD is then given by
\begin{equation}
    S(f)=|\tilde{Z}(f)|^2,
\end{equation}
where $\tilde{Z}(f)$ is the FFT of $Z(t)$ with $t=k\Delta t$. Finally, the spectrum is smoothed by calculating a running average with a range of 20 points.

\subsection{Simulation Parameters}

For all tasks we rescale the input and target sequences to the range between zero and one.
Before beginning the training and testing phases, the reservoirs are initialised by feeding in a sequence of 10000 inputs ${\bf i}_{in}(k)$. In the training phase 10000 inputs are used while in the testing phase 5000 inputs are used. 
The lengths of the training and testing phases were not optimised, as this was not the focus of this study. 

In the next section results are presented in dependence of the spectral radius of the initial reservoir $\rho_R$. The testing error measures are averaged over 100 realisations of the random input weights ${\bf W}_{in}$ and coupling weights ${\bf W}_{rand}$. For each $\rho_R$ value new random weights are drawn and the coupling matrix is rescaled such that the spectral radius is equal to the desired $\rho_R$. In each realisation the initial conditions for the Lorenz target trajectory are chosen randomly for the prediction phase, meaning that the start of the prediction phase is randomly distributed over the Lorenz attractor.

\section{Results and Discussion}\label{Sec:Results}

\begin{figure}[t]
\centerline{\includegraphics[width=0.45\textwidth]{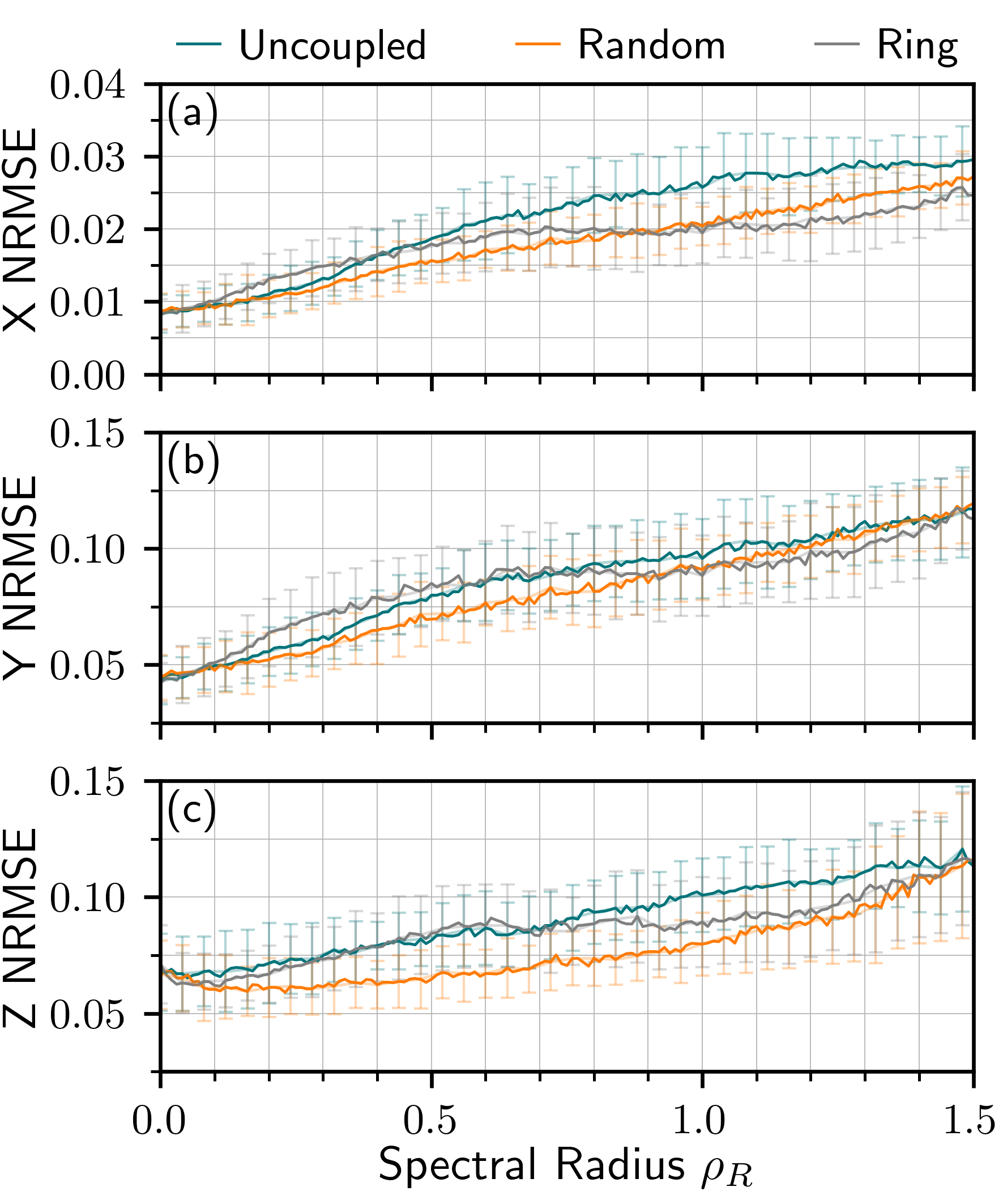}}
\caption{NRMSE of the $X$, $Y$ and $Z$ components for one-step-ahead prediction in open-loop operation as a function of spectral radius $\rho$ of the coupling matrix. The mean over 100 realisations of the random weights and Lorenz trajectories is depicted for the uncoupled (teal), random (orange) and ring (grey) reservoirs. The error bars indicate the standard deviation.}
\label{fig:nrmse}
\end{figure}

To judge the forecasting capabilities of a reservoir, typically only the performance in closed-loop operation is considered. We take a different approach, and assess the quality of the one-step-ahead by calculating the NRMSE on a test timeseries of 5000 steps in open-loop operation, i.e. the true Lorenz timeseries is used as the input throughout the testing phase. We do this to gain understanding about the influence of the training accuracy on the closed-loop prediction performance. Figure.~\ref{fig:nrmse} shows the NRMSE for the $X$, $Y$ and $Z$ components of the Lorenz timeseries for the three reservoir topologies. As a function of the spectral radius of the coupling martices, $\rho_R$, the NRMSEs are very similar for all three topologies, with the random network producing slightly lower errors. The lowest errors are achieved near $\rho_R=0$, which is the ELM limit. For $\rho_R=0$ the reservoir has no memory of past inputs, meaning that the predictions are based entirely on the current input step. This lack of memory is not detrimental to the task performance since the state of the Lorenz system at any point in time is fully defined by the current $X$, $Y$ and $Z$ values and these are all being provided as input. A study of the difference in reservoir performance for partial and full input information in given in \cite{STO22}.

\begin{figure}[t]
\centerline{\includegraphics[width=0.45\textwidth]{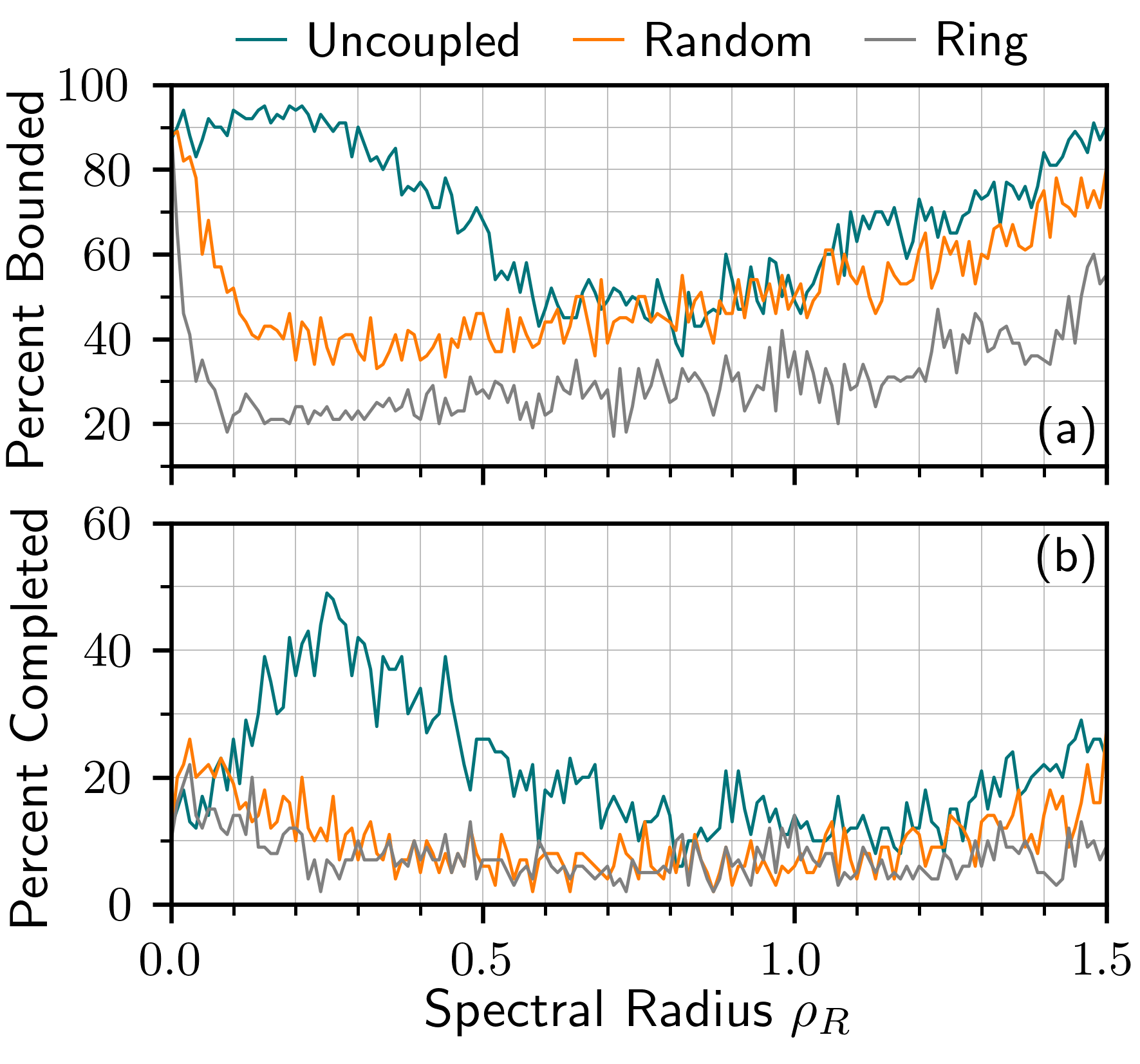}}
\caption{Percentage of the 100 realisations of the random weights and Lorenz trajectories for which the resulting closed-loop (autonomous) trajectory; (a) was bounded by $|X|,|Y|,|Z|<2$ for the entire prediction phase of 5000 steps, (b) was bounded and remained oscillatory for the entire prediction phase of 5000 steps.}
\label{fig:bounded}
\end{figure}

In the closed-loop (autonomous) operation, the predicted $X$, $Y$ and $Z$ values are fed back into the reservoir as the input. In this case three types of resulting dynamics are found; diverging, oscillatory and fixed point. In the diverging case the dynamic variables become much larger than the target values and therefore the attractor is not successfully reconstructed. Since we are interested in accurate timeseries prediction and attractor reconstruction, it is not relevant in this study if the dynamics are truely diverging or if the resulting system produces large oscillations. We therefore set an upper limit of two and check if $|X|$, $|Y|$ and $|Z|$ remain below this value for the entire prediction phase. Trajectories that remain below this limit we refer to as bounded. In Fig.~\ref{fig:bounded}a the percentage of the realisations which were bounded is depicted as a function of the spectral radius. Here, a clear difference can be seen between the uncoupled reservoir and the random and ring topologies. The uncoupled reservoir produces near 90\% bounded trajectories over a much large range of $\rho_R$. The bounded results can be separated into those that remain oscillatory for the entire prediction phase and those that collapse into a fixed point. The percentage of the results which remain oscillatory for the complete prediction phase is shown in Fig.~\ref{fig:bounded}b. Here, a clear difference is also evident between uncoupled reservoir and the random and ring topologies. Figure~\ref{fig:bounded} shows that the uncoupled reservoir produces bounded results more reliably and that fewer of the results collapse into a fixed point.

\begin{figure}[t]
\centerline{\includegraphics[width=0.9\textwidth]{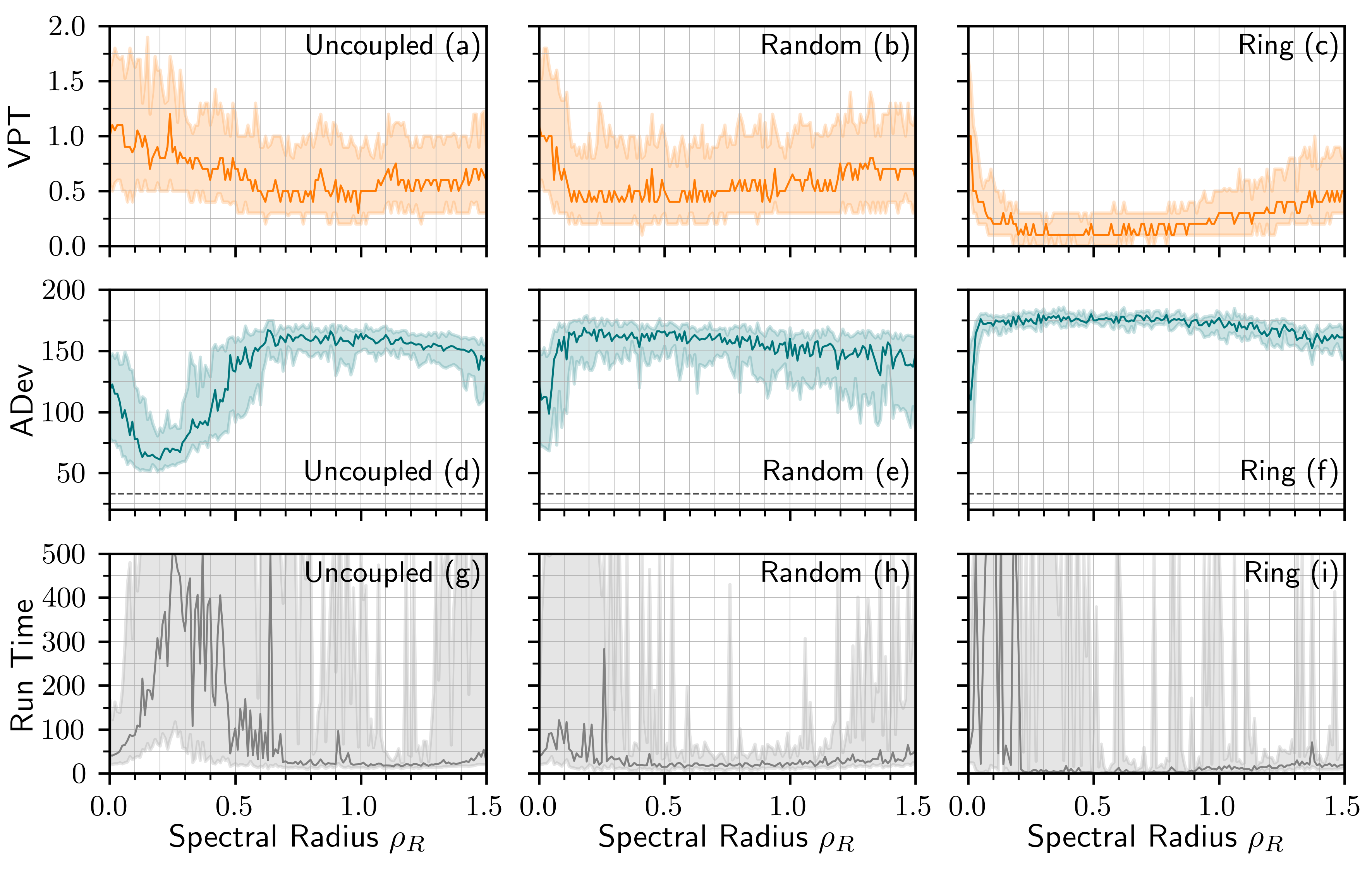}}
\caption{Closed-loop prediction performance: (a)-(c) The VPT for the three reservoir topologies. (d)-(f) The attractor deviation ADev, with a reference ADev calculated for two true Lorenz trajectories indicated by the dashed line. (g)-(i) The time interval over which 
 the trajectories were oscillatory. For the ADev and run time values only the bounded trajectories were considered. For all plots the median is plotted and the shaded regions show the 25$^\textrm{th}$ to 75$^\textrm{th}$ percentiles.}
\label{fig:VPT}
\end{figure}

To evaluate the short term closed-loop prediction performance the valid prediction time is plotted in Fig.~\ref{fig:VPT}a-c for the uncoupled, random and ring topologies, respectively. For $\rho_R=0$ all three systems are identical, and consequently show very similar performance as $\rho_R$ tends to zero. The best performance in terms of the VPT is achieved near $\rho_R=0$ for all three topologies. This correlates with the $\rho_R$ values for which the lowest NRMSE errors are achieved in the open-loop operations, see Fig.~\ref{fig:nrmse}. But, in the case of the uncoupled reservoir, this does not correspond to the $\rho_R$ values for which bounded and oscillatory trajectories are produced most reliably, see Fig.~\ref{fig:bounded}.
The finding that the highest VPT of to the Lorenz task are achieved at $\rho_R=0$ are in agreement with the results presented in \cite{VIE23}. In \cite{VIE23} reservoir hyperparameters are optimised for various tasks and for the Lorenz task it is shown that the best prediction performance is achieved when the adjacency matrix density is zero, which is equivalent to $\rho_R=0$ in this study.

\begin{figure}[t]
\centerline{\includegraphics[width=0.45\textwidth]{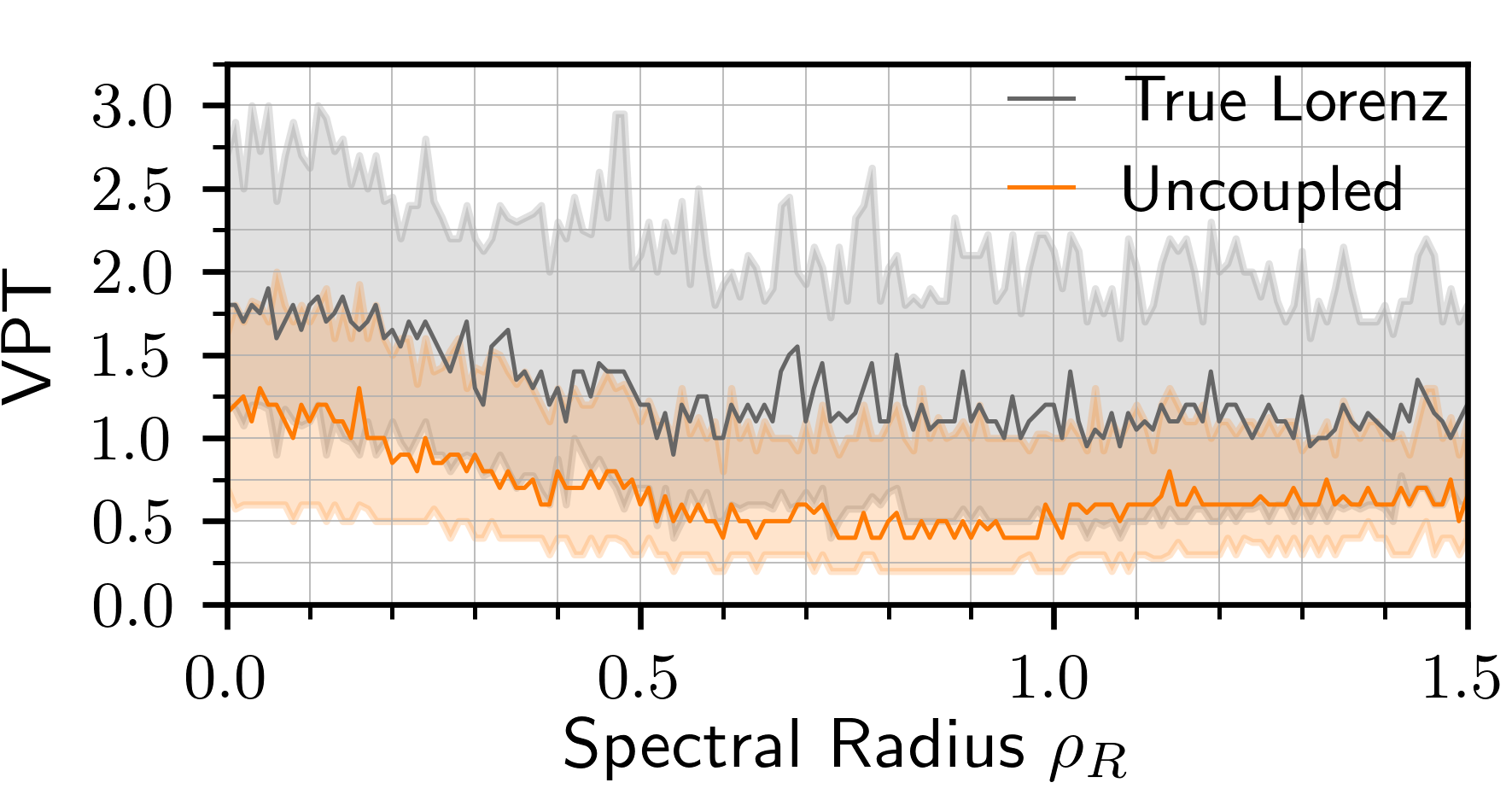}}
\caption{VPT for the uncoupled reservoir (orange) and for the true Lorenz system (grey). For the true Lorenz VPTs the first prediction step of the trained reservoir was used as the initial conditions the Lorenz equations and these were then integrated to generate the predicted timeseries. The median is plotted and the shaded regions show the 25$^\textrm{th}$ to 75$^\textrm{th}$ percentiles.}
\label{fig:VPT2}
\end{figure}

To judge the absolute performance it is useful to compare with other studies. 
However, when making such comparisons differences, in the task implementation and the error measures must be considered. In \cite{KOE23} the Lorenz task is implemented with the same time discretisation as in this study and the same definition for the valid prediction time is used. There VPTs of approximately 2 are achieved using a network of 100 nodes. By coupling the reservoir with a SINDy model, this is improved to VPTs of about 3. In \cite{PAT18a} a hybrid reservoir of 50 nodes coupled with a knowledge based model, resulted in VPTs of appproximately 1.5. The VPTs in \cite{KOE23} and \cite{PAT18a} are larger than in this study, however this was only achieved with larger node numbers and hybrid reservoir methods.

As well as information about the performance relative to other reservoir computing or machine learning approaches, we are also interested in the performance relative to the Lorenz system itself. The largest Lyaponov exponent of the Lorenz system indicates how quickly two trajectories on the true attractor will diverge given a certain difference in the initial conditions. Therefore, to obtain an objective benchmark with which the absolute performance in terms of the VPT can be compared, we calculate the VPT of the true Lorenz system given the error after the first prediction step (one prediction step is made using the trained reservoir, then this prediction is used as the initial conditions for the Lorenz system to generate a second timeseries with which to compare the target trajectory). This comparison is useful as it sets an upper limit on the VPT that can be expected given the accuracy in the one-step-ahead training stage. This comparison is depicted in Fig.~\ref{fig:VPT2} for the uncoupled reservoir. As is to be expected, the VPTs of the true Lorenz system are slightly larger, however the 75$^{\textrm{th}}$ percentile of the reservoir results coincide with the median of the true Lorenz VPTs, indicating considerable overlap in the performance.

\begin{figure}[t]
\centerline{\includegraphics[width=0.9\textwidth]{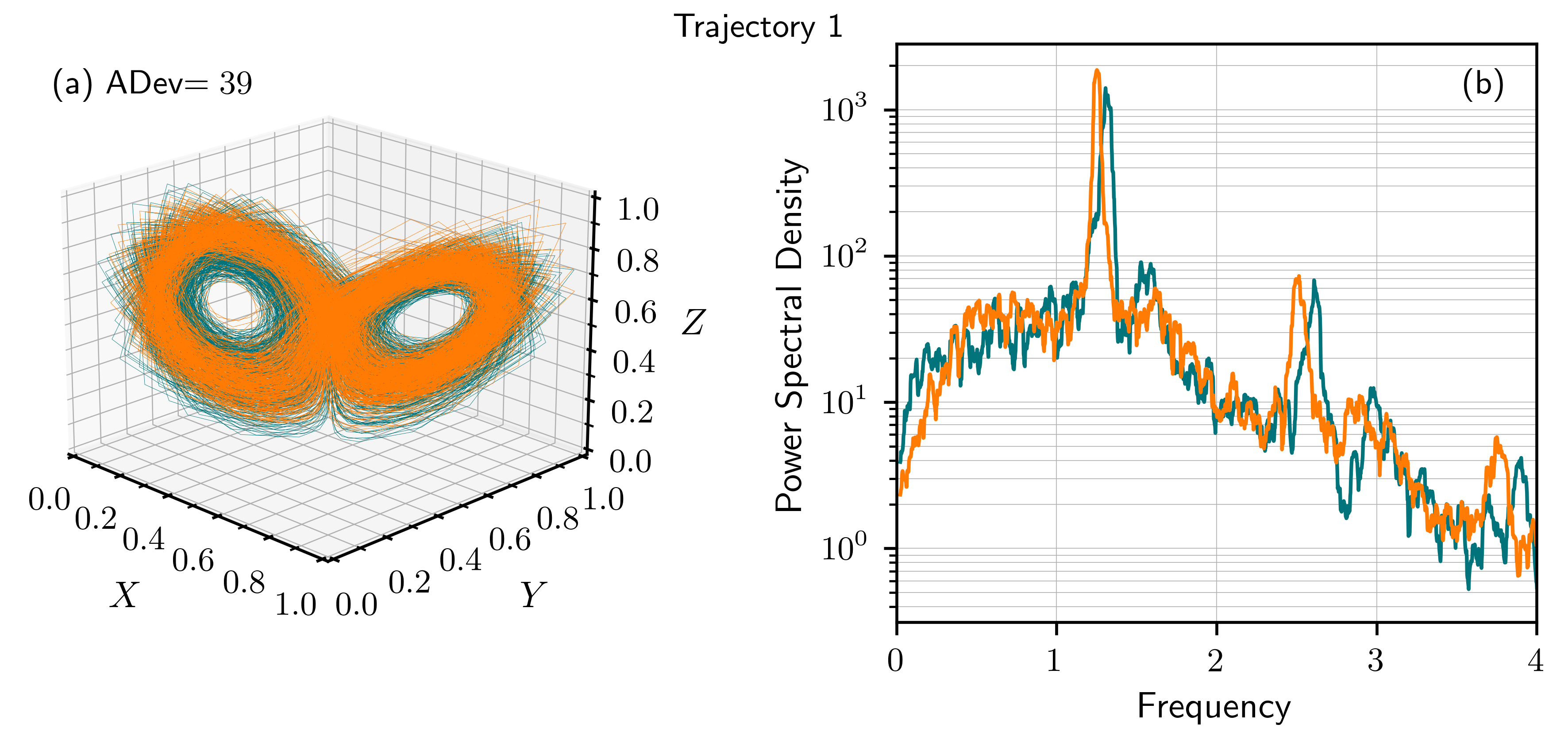}}
\centerline{\includegraphics[width=0.9\textwidth]{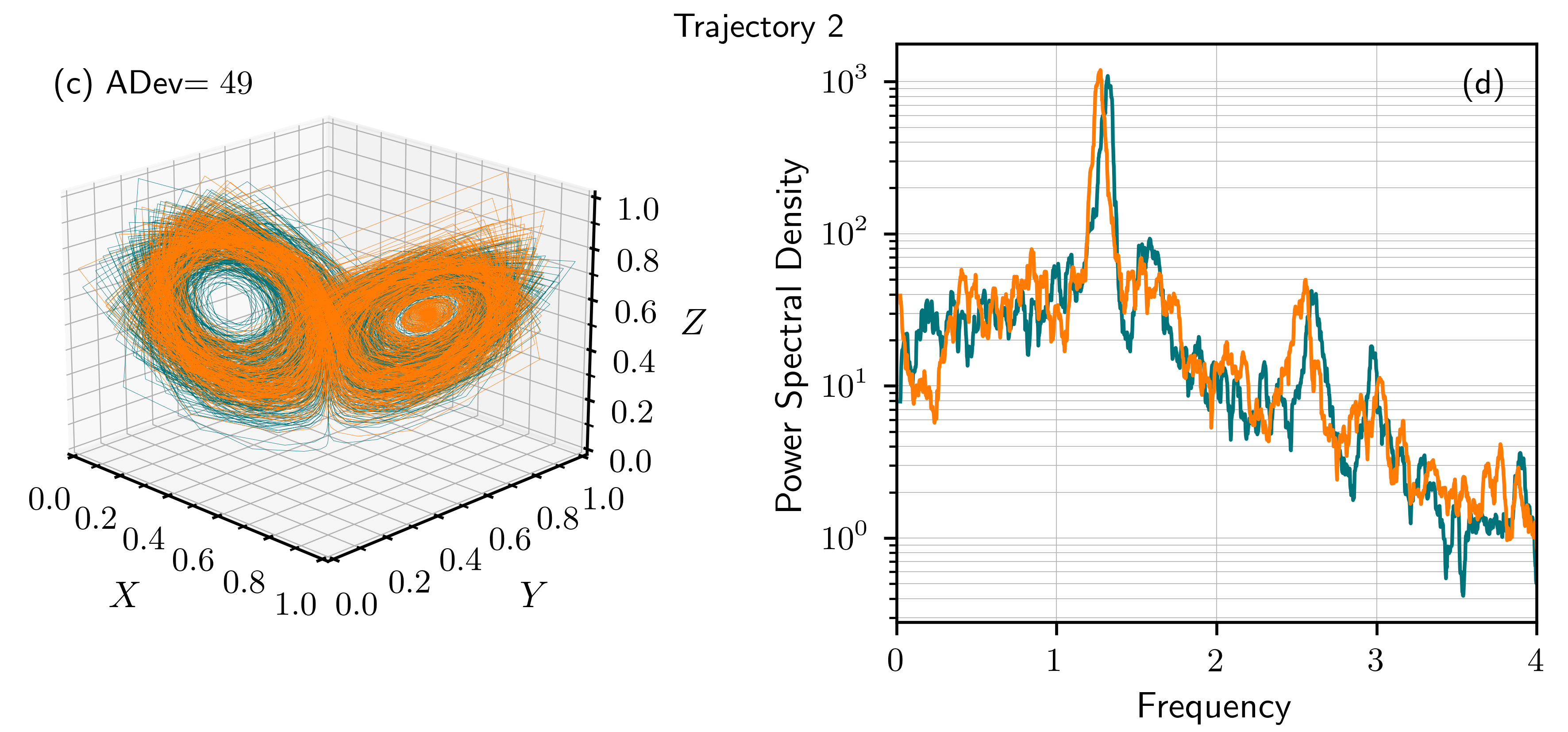}}
\centerline{\includegraphics[width=0.9\textwidth]{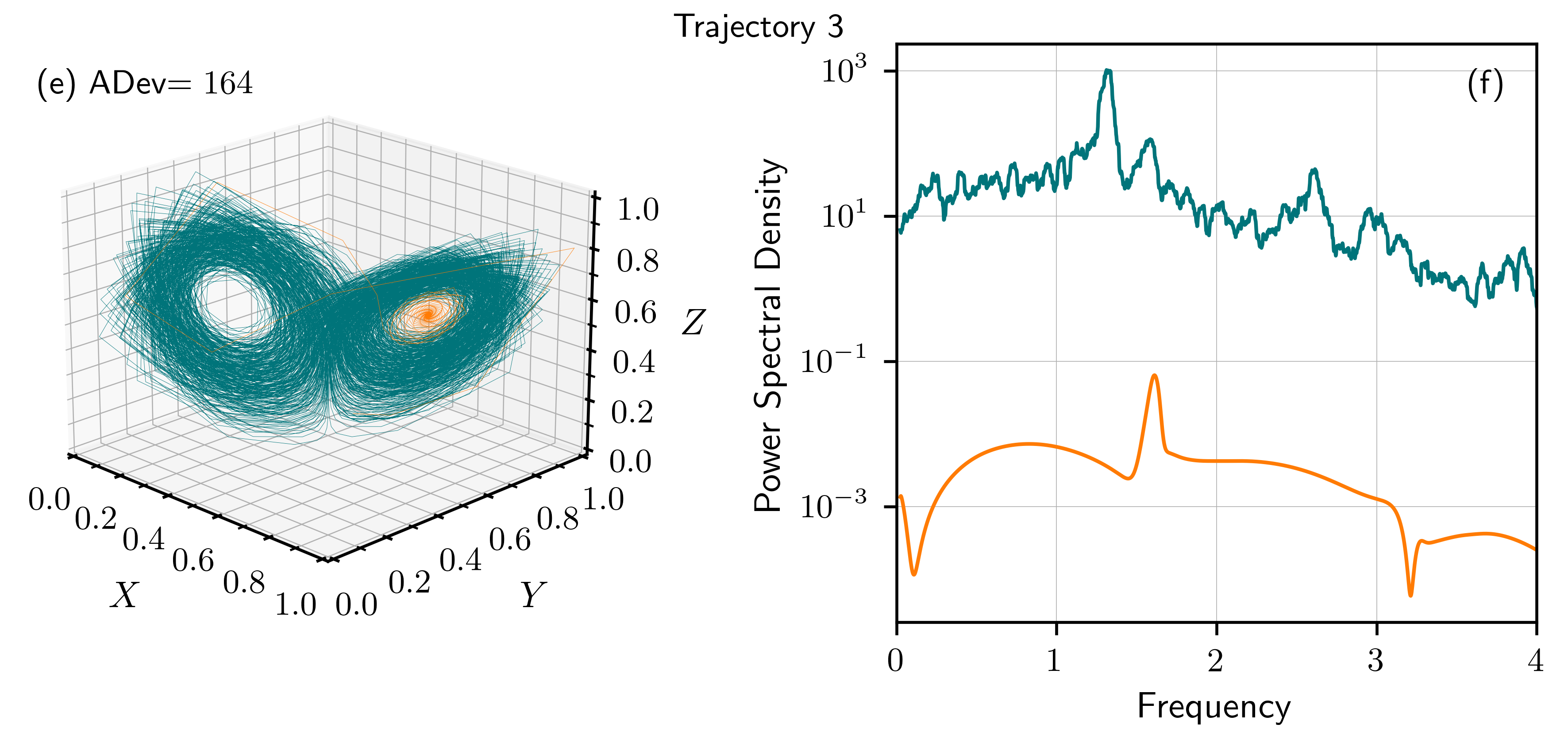}}
\caption{Phase spaces plots (a,c,e) and the corresponding PSD plots (b,d,f) for three example trajectories generated using the uncoupled reservoir in closed-loop operation. The predicted results are shown in orange and the target results in green. The trajectories in (c,d) and (e,f) correspond to trajectories two and three in Fig.~\ref{fig:timeseries}, respectively.}
\label{fig:fft}
\end{figure}

Figure~\ref{fig:VPT2} shows that even for the true Lorenz system, given differences in the initial conditions of the order of the NRMSEs depicted in Fig.~\ref{fig:nrmse}, trajectories diverge after one to three Lyaponov times ($1/\lambda_L\approx 1.1$). Therefore, to evaluate the quality of the attractor reconstruction over hundreds of Lyaponov times we use the attractor deviation measure introduced in Section~\ref{sec:adev}. The attractor deviation (ADev) is a measure of the difference in the 3-dimensional phase space which is traversed by the trajectory of the target Lorenz timeseries and the predicted timeseries. Figure~\ref{fig:VPT}d-f shows the attractor deviation for the uncoupled, random and ring reservoirs. Only the realisations resulting in bounded trajectories were considered. The horizontal dashed line is a reference attractor deviation of two true Lorenz trajectories. The random and ring topologies show very similar performance, and their large attractor deviations can be explained by the low percentage of trajectories which do not collapse into a fixed point during the prediction phase (as shown in Fig.~\ref{fig:bounded}b). The uncoupled reservoir, on the other hand, shows quite low attractor deviation values corresponding to the $\rho_R$ range for which a high percentage of the trajectories remain oscillatory for the entire prediction phase.

\begin{figure}[t]
\centerline{\includegraphics[width=0.9\textwidth]{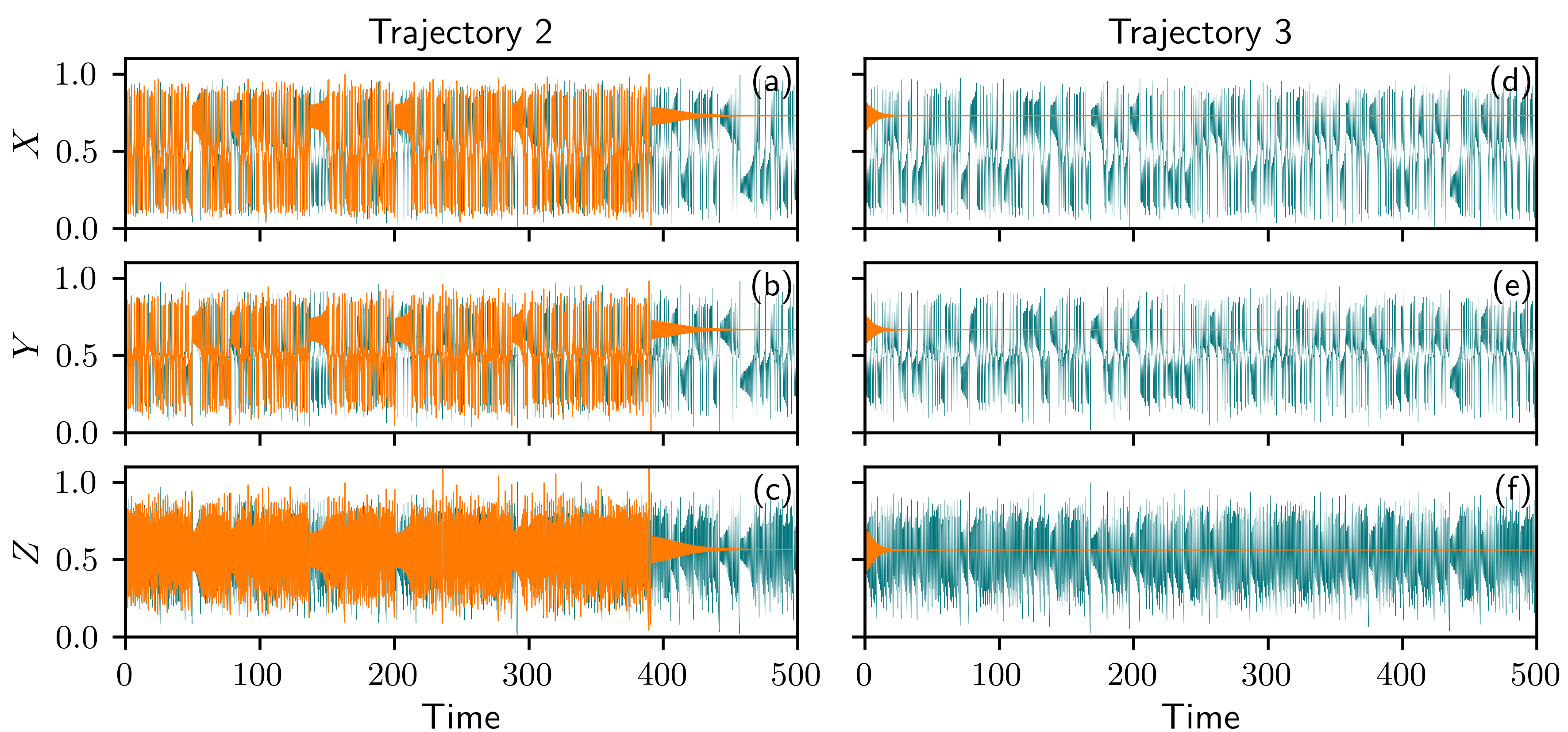}}
\caption{Timeseries of the $X$, $Y$ and $Z$ components for two predicted trajectories generated using the uncoupled reservoir in closed-loop operation. The predicted results are shown in orange and the target results in green. Trajectories two and three correspond to the trajectories in (c,d) and (e,f)  of Fig.~\ref{fig:fft}, respectively.}
\label{fig:timeseries}
\end{figure}

Since the attractor deviation measure does not capture any temporal information, it is also useful to compare the power spectral densities of the predicted and true trajectories. Figure~\ref{fig:fft} shows phase-space plots, power spectral densities and the attractor deviation for three example trajectories generated using the uncoupled reservoir. Different random input weights and initial conditions of the Lorenz system were used to generate the trajectories. The trajectory in Fig.~\ref{fig:fft}b,c collapses near the end of the prediction phase and corresponds to trajectory two in Fig.~\ref{fig:timeseries}. Figure~\ref{fig:fft}e,f corresponds to trajectory three in Fig.~\ref{fig:timeseries}. As long as the trajectory remains on the attractor, the spectral features are reproduced quite well and are comparable with the performance achieved using a data-informed-reservoir hybrid model in \cite{KOE23}, even though a much smaller reservoir is used in this study.

The Lorenz-like dynamics achieved with the trained reservoir appear to be meta-stable. Meaning that the trajectories can appear Lorenz like for hundreds of Lyaponov times and then suddenly collapse to a stable fixed-point, as depicted in the timeseries shown in Fig.~\ref{fig:timeseries}. Similiar behaviour has also been reported in other works \cite{HAL19}. 
Comparing the results depicted in Fig.~\ref{fig:nrmse}, Fig.~\ref{fig:bounded} and Fig.~\ref{fig:VPT} a trade-off between the short-term prediction performance and the stable attractor reconstruction is evident.

\begin{figure}[t]
\centerline{\includegraphics[width=0.9\textwidth]{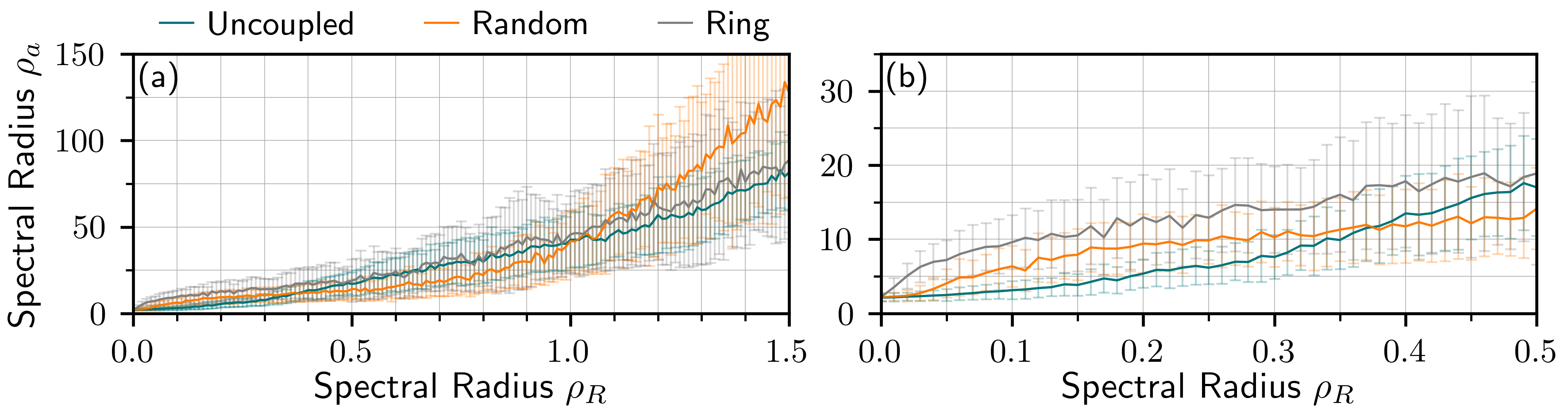}}
\caption{ (a) Spectral radius of the trained autonomous system $\rho_a$ as a function of the spectral radius of the original reservoir $\rho$. The solid lines indicate the mean and the error bars give the standard deviation. (b) Enlargement of the $\rho_R\in\left[0,0.5\right]$ range of (a).}
\label{fig:rho}
\end{figure}

For the reservoir to accurately reconstruct the Lorenz attractor, the Lorenz-like dynamics of the trained system must be stable. For the uncoupled reservoir a large percentage of the solutions appear to be stable or meta-stable over the prediction range. However, for the random and ring topologies most of the solutions diverge quickly, indicating that the desired solutions either do not exist in the final autonomous system or are unstable. To work towards understanding the difference in the results between the reservoir topologies, the spectral radius of the trained autonomous systems $\rho_a$ is plotted in Fig.~\ref{fig:rho} as a function of the spectral radius of the original reservoir. For all three reservoirs $\rho_a$ increases with $\rho_R$ and for $\rho_R>0.5$ the large $\rho_a$ values correspond to short run times (time before the system diverges or collapses to a fixed point), as shown in Fig.~\ref{fig:VPT}g-i. For $\rho_R<0.35$, $\rho_a$ of the uncoupled system is smaller than those of the random and ring reservoirs (see Fig.~\ref{fig:rho}b for an enlargement of the $\rho_R<0.5$ results). The region of low $\rho_a$ for the uncoupled system corresponds to the regions of lower attractor deviation and longer run times in Fig.~\ref{fig:VPT}d,g.
These results indicate that the topology of the initial reservoir plays a deciding role in determining the stability of the desired solution in the final system. Specifically, less connectivity within the network leads to improved results. These results are in agreement with the findings of \cite{MA23} where the authors showed improved performance for block diagonal coupling matrices with low node degree (zero node degree was not considered).

\section{Conclusion}\label{Sec:Conclusion}

In this work the chaotic Lorenz attractor has been reconstructed using small reservoirs of only 20 nodes. We have shown that the short term performance depends on the error in the one-step-ahead prediction, but that the stable reconstruction of the attractor is related to dynamical properties of the trained autonomous system. By reducing the reservoir to a system of uncoupled nodes, smaller spectral radii of the trained system can be achieved compared with random or ring reservoirs. And, the parameter regions resulting in small spectral radii for the trained uncoupled reservoir correspond to more reliable reconstruction of the Lorenz attractor.

These results pave the way towards interpretable surrogate models. Due to the small size of the reservoir and the uncoupled reservoir nodes it becomes feasible to perform bifurcation analysis on the final system and additional optimisation can be performed after the training phase without effecting all reservoir nodes, for example pruning of selected nodes. Furthermore, hardware implementability is improved, as complex network topologies are not required and for a system of uncoupled nodes there is more freedom in the choice of multiplexing. 

The insights gained in this study into the relationships between training accuracy, network topology and short- and long-term prediction performance can also be applied to related machine learning approaches. For example, nonlinear vector regression (referred to as ”next generation reservoir computing” \cite{GAU21b}) has recently been shown to perform very well at chaotic timeseries forecasting. Nonlinear vector regression could be thought of a network of node degree zero (and no self-coupling) and would then be similar to the uncoupled reservoir in this study. 

In this study all three dynamical variables of the Lorenz system were provided to the reservoir in the training stage. This meant that the reservoir did not need to have memory for the one-step-ahead training to perform well. In contrast, for tasks where only partial information is provided in the training stage the reservoir must perform a delay embedding and therefore requires memory \cite{LUK09,HAR20b,STO22}. Since the memory of the reservoir is determined by the coupling topology and the spectral radius, different dependencies on these properties can be excepted for partial information tasks compared with the results for the full information Lorenz task presented in this study. In order to restrict the reservoir topology to uncoupled nodes for partial information tasks, but still incorporate the required memory, a possible solution could be to apply memory augmentation methods at the levels of the input or output layers. In \cite{JAU24} it was shown that adding task relevant timescales at the level of the input or output layers can improve the reservoir performance for partial information time series prediction tasks and decreases the sensitivity to other reservoir hyperparameters.

\clearpage

\section*{References}
\providecommand{\newblock}{}

\end{document}